\documentclass[sigconf]{acmart}

\AtBeginDocument{%
  }

\usepackage{multirow}
\usepackage{url}

\usepackage{pifont}
\usepackage{makecell}
\usepackage{hyperref}
\usepackage{acmart-taps}

\begin{document}

\title{Gesture Generation from Trimodal Context for Humanoid Robots}


\author{Shiyi Tang}
\affiliation{%
  \institution{Heriot-Watt University}
\country{UK}
}
\email{st2015@hw.ac.uk}

\author{Christian Dondrup}
\affiliation{%
  \institution{Heriot-Watt University}
\country{UK}
}
\email{c.dondrup@hw.ac.uk}







\begin{abstract}
 Natural co-speech gestures are essential components to improve the experience of Human-robot interaction (HRI). However, current gesture generation approaches have many limitations of not being natural, not aligning with the speech and content, or the lack of diverse speaker styles. Therefore, this work aims to repoduce the work by \cite{27} generating natural gestures in simulation based on tri-modal inputs and apply this to a robot. During evaluation, ``motion variance'' and ``Frechet Gesture Distance (FGD)'' is employed to evaluate the performance objectively. Then, human participants were recruited to subjectively evaluate the gestures. Results show that the movements in that paper have been successfully transferred to the robot and the gestures have diverse styles and are correlated with the speech. Moreover, there is a significant likeability and style difference between different gestures.
\end{abstract}





\maketitle

\section{Introduction}\label{sec:introdcution}
Gestures are non-linguistic and can enhance communication when combined with speech \cite{1}. However, generating natural and diverse gestures is challenging \cite{6} and issues of lack of style, unnaturalness, and poor alignment with speech context persist.

\begin{figure}[h]
    \centering
        \includegraphics[width=0.5\linewidth]{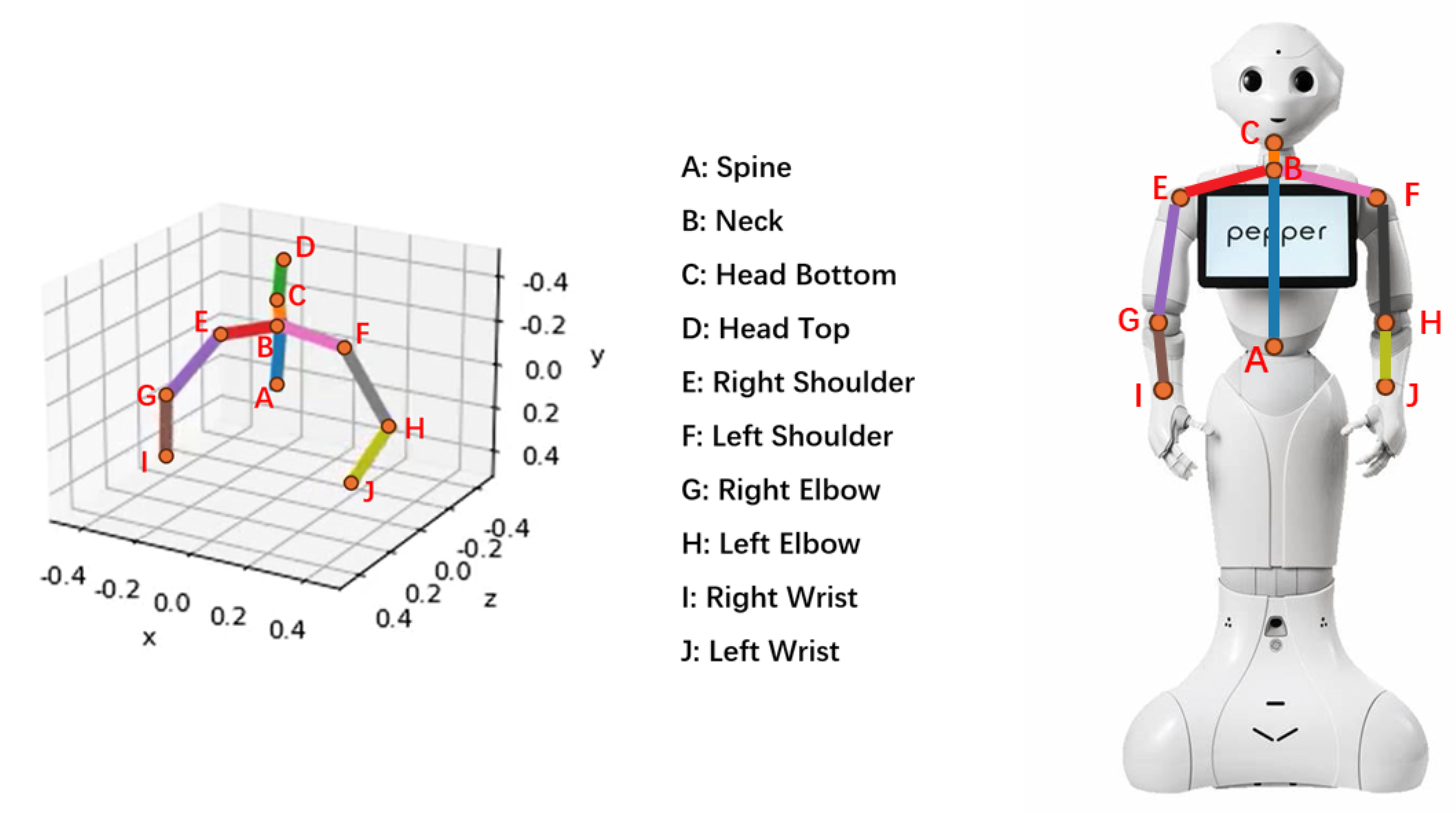}
        \caption{Transformation between the stick figure and the robot. Image of Pepper taken from ~\cite{picture}.}
        \label{fig:transformation}
    \end{figure}

Yoon et al. introduced an end-to-end gesture generation framework with trimodal input \cite{27}. This model outperforms previous end-to-end models and can generate different gesture styles (e.g., introverted or extroverted) for the same sentence due to the speaker identity \cite{27}. However, the viability of this approach for conversion of the movements of human to a robot with fewer Degrees of Freedom (DoF) (Fig~\ref{fig:transformation}) was not proved, nor did they separately evaluate motion quality and diversity in their user study.

This paper aims to reproduce Yoon et al.'s work \cite{27}, apply it to Pepper~\cite{pepper}, and extend the user study. Results are applied to Pepper using Kinematics with additional angle and velocity adjustments not previously done. Also, likeability and speech-gesture correlation between different gesture styles, and the performance of the originally generated gesture and the robot gesture are compared in detail.

    \begin{figure}[h]
        \centering
        \includegraphics[width=0.3\linewidth]{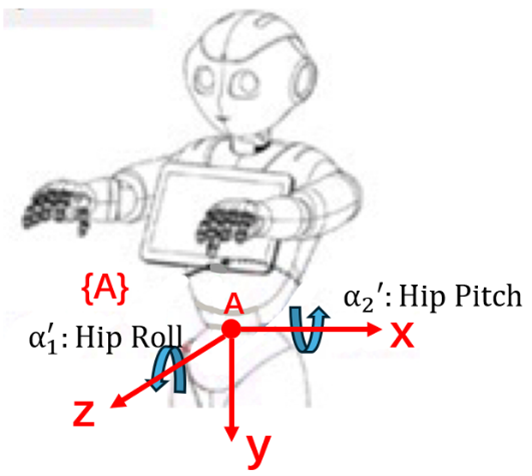}
        \caption{Robot coordinate of Hip.}
        \label{fig:hip-cor}
    \end{figure}


\section{Methodology}\label{sec:methodology}
Initially, Google TTS synthesises speech audio from custom input text. Then, the audio, text, and speaker ID are inputs to the \textbf{Pose Generation} module and produce 3D poses frame by frame, with each pose containing 3D coordinates for 10 joints and are visualised as stick figures in Fig.~\ref{fig:transformation} \footnote{Images of pepper's body shown in the following sections are taken from  (http://doc.aldebaran.com/2-5/family/pepper\_technical/joints\_pep.html)}. Then, the \textbf{Pose2Angle} module calculates rotation angles. For example, $\alpha_1$ and $\alpha_2$ are HipRoll and HipPitch angles. The robot coordinate {A} is built in Fig \ref{fig:hip-cor}, where $\alpha_1{ }^{\prime}$ and $\alpha_2{ }^{\prime}$ in range of [$-\pi$, $\pi$] are the rotation angles of $\overrightarrow{AB}$ (Fig ~\ref{fig:hip-1} and Fig ~\ref{fig:hip-2}). Moreover, 2 constant values \( m \) and \( n \) are introduced and allow to manually adjust the performance of the robot. Given 3D coordinates of A and B: $(A x, A y, A z)$ and $(B x, B y, B z)$, $$\left\{\begin{array}{l}\alpha_1=\left\{\begin{array}{l}\left(\alpha_1^{\prime}+\pi\right) * m, \quad \text {if} \quad \alpha_1^{\prime}<0 \\ \left(\alpha_1^{\prime}-\pi\right) * m, \quad \text {if} \quad \alpha_1^{\prime}>0\end{array} \quad m=0.3\right. \\ \alpha_2=\left\{\begin{array}{l}-\left(\alpha_2^{\prime}+\pi\right) * n, \quad \text {if} \quad \alpha_2^{\prime}<0 \\ -\left(\alpha_2^{\prime}-\pi\right) * n, \quad\text {if} \quad \alpha_2^{\prime}>0\end{array} \quad n=0.3\right.\end{array}\right.$$, where $\left\{\begin{array}{l}\alpha_1^{\prime}=\operatorname{atan2}\left(B_x-A_x, B_y-A_y\right) \\ \alpha_2^{\prime}=\operatorname{atan2}\left(B_z-A_z, B_y-A_y\right)\end{array}\right.$.

To prevent the velocity from exceeding the robot's joint limits, velocities of the next time step are adjusted. $\theta_i$ and $\theta_{i+1}$ are the rotation angle of a joint at time $i$ and $i+1$. The adjusted angle $\left\{\begin{array}{l}\theta_{i+1}^{\prime}=\boldsymbol{v e l}_{\text {max }} * \boldsymbol{t}+\boldsymbol{\theta}_{\boldsymbol{i}}, \quad i f \quad \theta_{i+1}>\theta_i \\ \theta_{i+1}^{\prime}=-vel_{\max } * \boldsymbol{t}+\boldsymbol{\theta}_{\boldsymbol{i}}, \quad if \quad \theta_{i+1}<\theta_i\end{array}\right.$. Finally, the \textbf{PlayGesture} module uses Naoqi's python API to enable the Pepper robot to play the audio while performing the gestures.

\aptLtoX[graphic=no,type=html]{
\begin{figure}[htbp]
    \centering
        \includegraphics[width=0.6\linewidth]{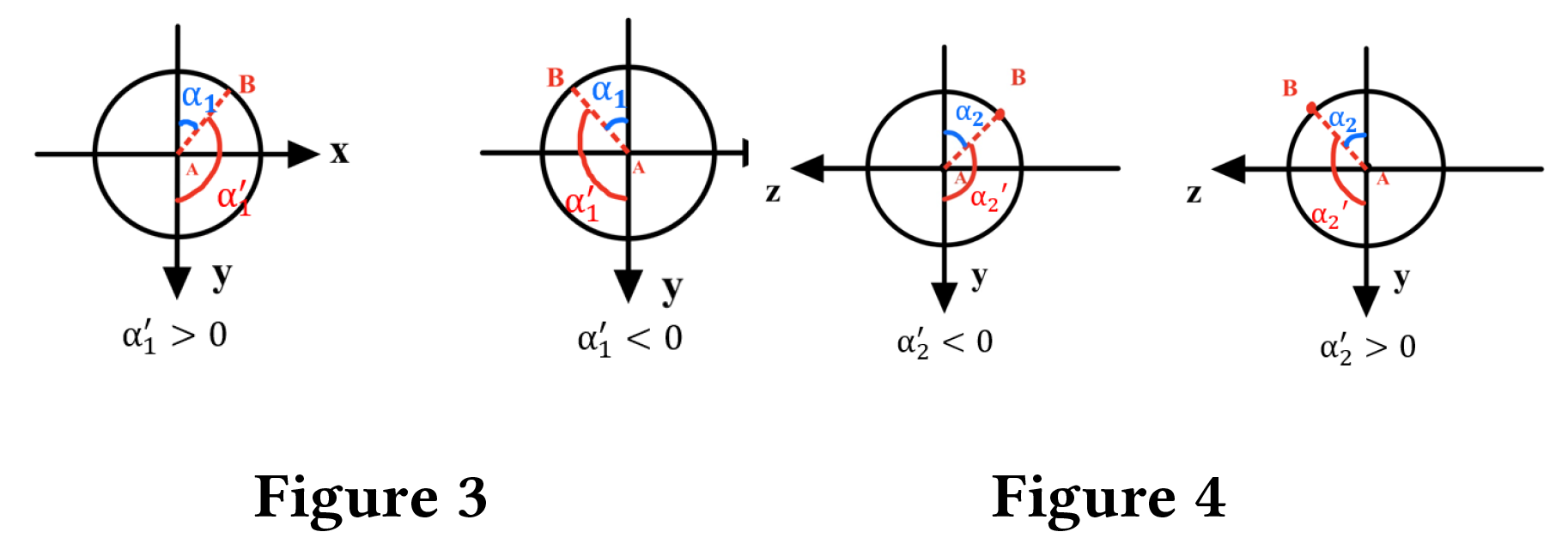}
        \caption{}
        \label{fig:hip-1}
        \label{fig:hip-2}
\end{figure}
}{\begin{figure}[htbp]
    \centering
    \begin{minipage}[t]{0.2\textwidth}
        \centering
        \includegraphics[width=4cm]{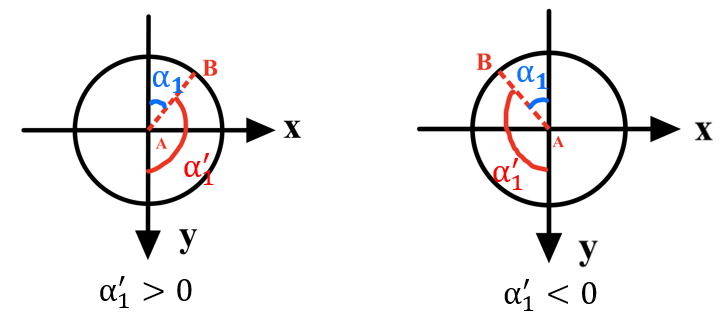}
        \caption[]{}
        \label{fig:hip-1}
    \end{minipage}
    \begin{minipage}[t]{0.2\textwidth}
        \centering
        \includegraphics[width=4cm]{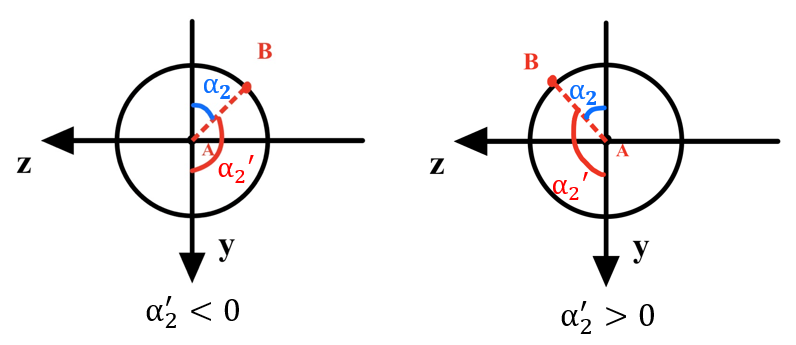}
        \caption[]{}
        \label{fig:hip-2}
    \end{minipage}
\end{figure}}

\section{Experimental Design and Results}\label{sec:results}
The research questions are as follows: \textbf{RQ1:} What are the differences in gesture movement performance between the robot and the stick figure? \textbf{RQ2:} How will the gesture styles of the same input sentence differ given different speaker ids? \textbf{RQ3:} What is the difference in likeability between each gesture style? \textbf{RQ4:} Is there a correlation between speech and gesture?


    

    

    


According to Fig \ref{fig:distribution}, 3 speaker IDs were selected from \ding{192}, \ding{193}, and \ding{194} to represent introverted, normal, and extroverted styles. The $FGD$ (Fréchet distance \cite{27}) between extroverted and introverted styles is the largest (0.6274) compared to other 2 FGDs (0.3093 between extroverted and normal gestures, 0.4338 between introverted and normal gestures).

\setcounter{figure}{4}
\begin{figure}[ht]
    \centering
    \includegraphics[width=.5\linewidth]{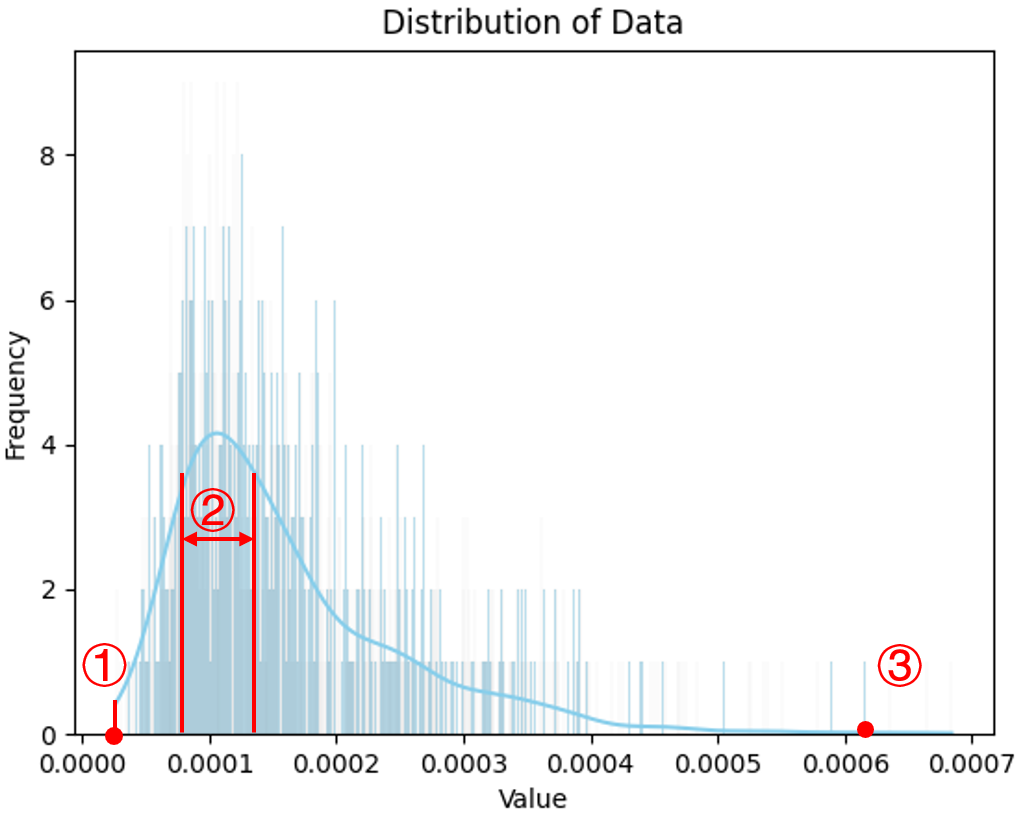}
    \caption[]{Distribution of the motion variance.}
    \label{fig:distribution}
\end{figure}

Then in subjective evaluation, after watching a video of the robot, 21 participants completed a questionnaire with the following indexes:  Anthropomorphism and Likeability from Godspeed, Speech Gesture Correlation ~\cite{27}, and Style (introverted - extroverted). 3 different sentences were randomly selected and a questionnaire for each of them was created. The number of answers for each sentence is 8, 7 and 6. Participants watched videos of both stick figure and robot with three different gesture styles for the same sentence. To avoid contrast and carryover effects, the videos were played separately in random order and participants evaluated only the movements.

There is no significant difference between the scores of the stick figure and the robot, indicating that the paper's results apply well to the robot and the velocity adjustment is not noticeable.


\begin{table}[htbp]
\centering
\scalebox{0.8}{\begin{tabular}{|l|l|l|l|}
\hline
Dependent variable                     & (I) Style          & (J) Style & p   \\
\hline
\multirow{2}{*}{Likeability}        & \multirow{2}{*}{1} & 2         & 0.0106 \\
\cline{3-4}
                                       &                    & 3         & 0.0028 \\
                                       \hline
\multirow{2}{*}{\makecell[c]{Speech Gesture Correlation}} & \multirow{2}{*}{1} & 2 & 0.038 \\
\cline{3-4}
                                       &                    & 3         & 0.017 \\
                                       \hline
\end{tabular}}
\caption{Post hoc test.}
\label{tab:post-hoc-likeCor}
\end{table}

\begin{figure}[ht]
    \centering
    \includegraphics[width=0.95\linewidth]{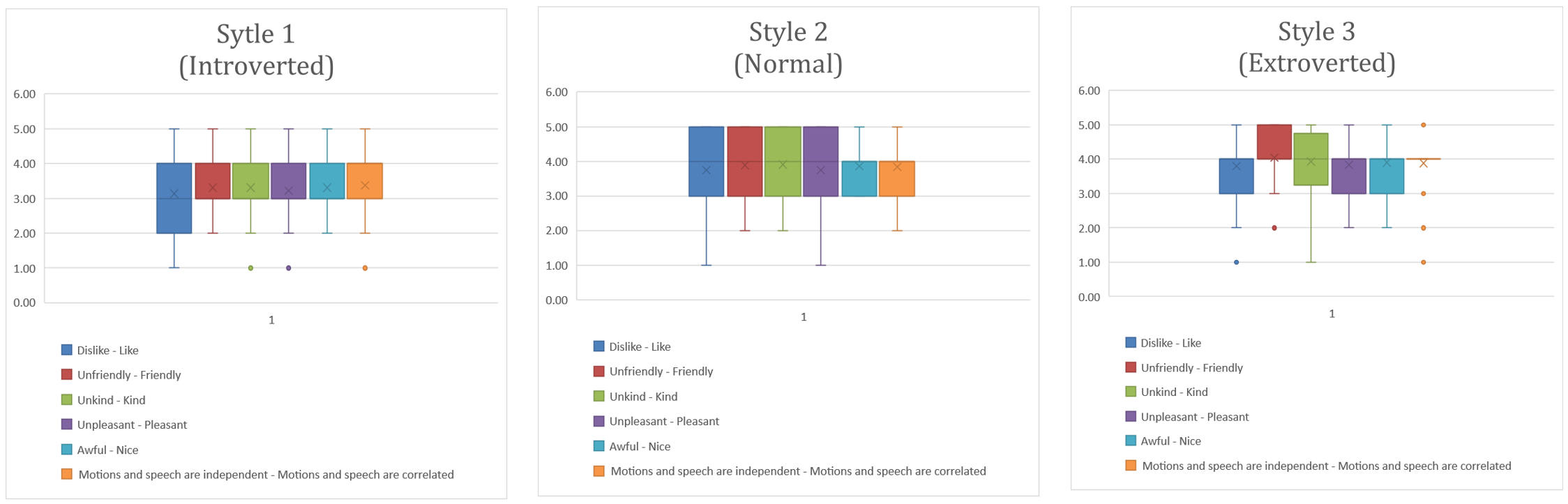}
    \caption[]{Comparison of different gesture styles.}
    \label{fig:style-dif}
\end{figure}

Also, the Likeability and Speech Gesture Correlation of style 1 (introverted) are significantly different from style 2 (normal) and 3 (extroverted) in Table \ref{tab:post-hoc-likeCor} and Fig \ref{fig:style-dif}. People prefer normal and extroverted gestures and perceive them as having a higher speech gesture correlation. Therefore, higher speech gesture correlation leads to greater likeability.

\begin{figure}[htbp]
    \centering
    \includegraphics[width=0.6\linewidth]{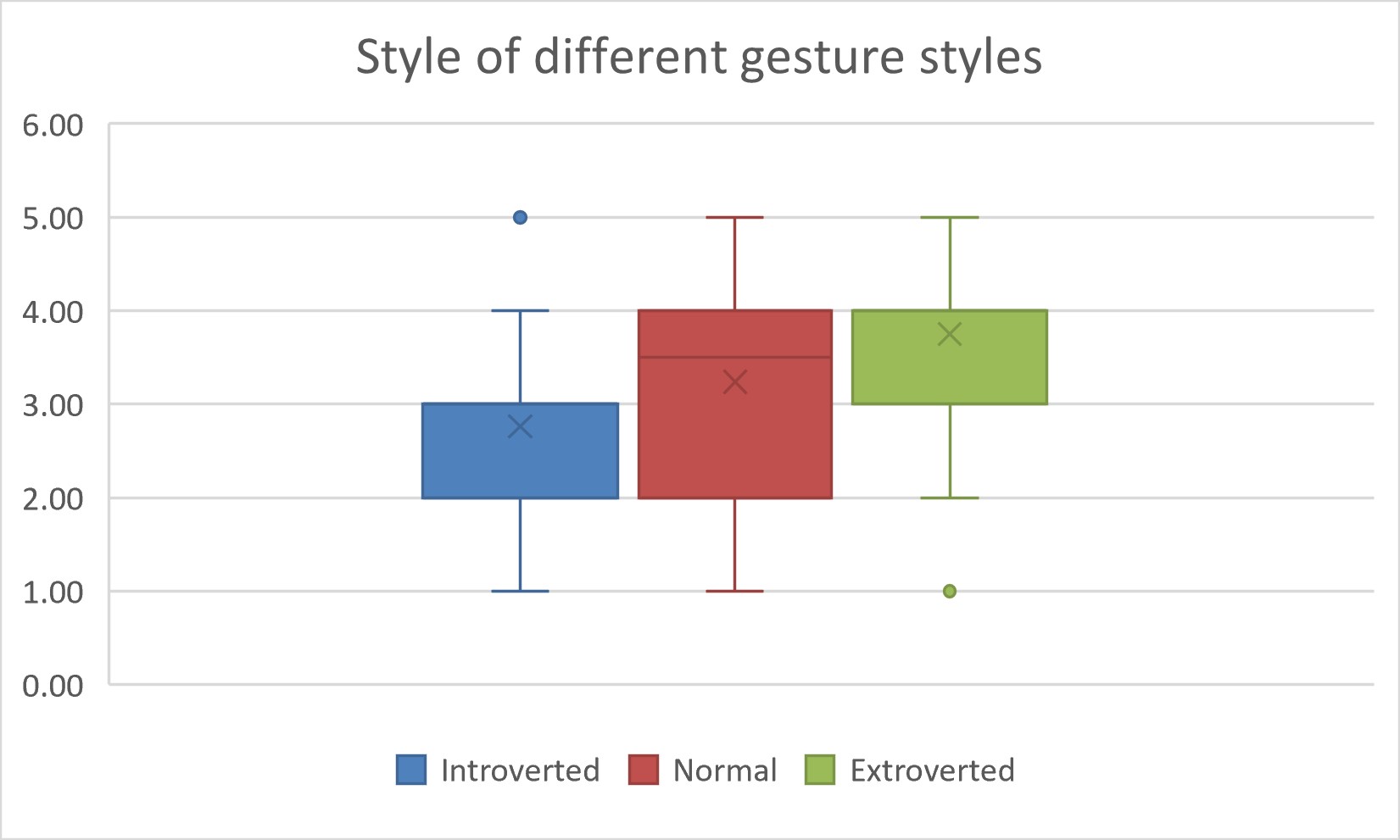}
    \caption[]{Style of different gesture styles.}
    \label{fig:styles}
\end{figure}

There is also a significant difference (p = 0.000) between the style of different gesture styles. Style 3 significantly differs from styles 1 (p = 0.000) and 2 (p = 0.032) as shown in Fig \ref{fig:styles}.

\section{Discussion}\label{sec:discussion}
Results showed no significant difference between the robot and the stick figure [RQ1], confirming successful movements transfer from the stick figure to the robot. Moreover, there is a clear sytle difference [RQ2] and people prefer extroverted and normal gestures over introverted ones [RQ3] and perceive extroverted ones has higher speech-gesture correlation [RQ4], which provides a direction to generate more likeable gestures. Future work will focus on an end-to-end model that directly outputs rotation angles within the robot's maximum velocity, which can eliminate the need for angle calculation and velocity adjustments. Future user studies will include between-subject experiments and recruit more participants. New research questions such as how different voice genders affect perception and likeability of gestures will also be explored.






\bibliographystyle{ACM-Reference-Format}
\bibliography{main}


\begin{thebibliography}{5}


\ifx \showCODEN    \undefined \def \showCODEN     #1{\unskip}     \fi
\ifx \showDOI      \undefined \def \showDOI       #1{#1}\fi
\ifx \showISBNx    \undefined \def \showISBNx     #1{\unskip}     \fi
\ifx \showISBNxiii \undefined \def \showISBNxiii  #1{\unskip}     \fi
\ifx \showISSN     \undefined \def \showISSN      #1{\unskip}     \fi
\ifx \showLCCN     \undefined \def \showLCCN      #1{\unskip}     \fi
\ifx \shownote     \undefined \def \shownote      #1{#1}          \fi
\ifx \showarticletitle \undefined \def \showarticletitle #1{#1}   \fi
\ifx \showURL      \undefined \def \showURL       {\relax}        \fi
\providecommand\bibfield[2]{#2}
\providecommand\bibinfo[2]{#2}
\providecommand\natexlab[1]{#1}
\providecommand\showeprint[2][]{arXiv:#2}

\bibitem[pep(2015)]%
        {pepper}
 \bibinfo{year}{2015}\natexlab{}.
\newblock \bibinfo{howpublished}{\url{https://www.aldebaran.com/en/pepper}}.
\newblock


\bibitem[pic(2015)]%
        {picture}
 \bibinfo{year}{2015}\natexlab{}.
\newblock \bibinfo{howpublished}{\url{https://www.gizlogic.com/wp-content/uploads/2015/06/Robot-Pepper.jpg}}.
\newblock


\bibitem[Goldin‐Meadow and McNeill(1999)]%
        {1}
\bibfield{author}{\bibinfo{person}{Susan Goldin‐Meadow} {and} \bibinfo{person}{David McNeill}.} \bibinfo{year}{1999}\natexlab{}.
\newblock \showarticletitle{The role of gesture and mimetic representation in making language the province of speech.}
\newblock
\urldef\tempurl%
\url{https://api.semanticscholar.org/CorpusID:151686384}
\showURL{%
\tempurl}


\bibitem[Nyatsanga et~al\mbox{.}(2023)]%
        {6}
\bibfield{author}{\bibinfo{person}{Simbarashe Nyatsanga}, \bibinfo{person}{Taras Kucherenko}, \bibinfo{person}{Chaitanya Ahuja}, \bibinfo{person}{Gustav~Eje Henter}, {and} \bibinfo{person}{Michael Neff}.} \bibinfo{year}{2023}\natexlab{}.
\newblock \showarticletitle{A Comprehensive Review of Data‐Driven Co‐Speech Gesture Generation}.
\newblock \bibinfo{journal}{\emph{Computer Graphics Forum}}  \bibinfo{volume}{42} (\bibinfo{year}{2023}).
\newblock
\urldef\tempurl%
\url{https://api.semanticscholar.org/CorpusID:255825797}
\showURL{%
\tempurl}


\bibitem[Yoon et~al\mbox{.}(2020)]%
        {27}
\bibfield{author}{\bibinfo{person}{Youngwoo Yoon}, \bibinfo{person}{Bok Cha}, \bibinfo{person}{Joo-Haeng Lee}, \bibinfo{person}{Minsu Jang}, \bibinfo{person}{Jaeyeon Lee}, \bibinfo{person}{Jaehong Kim}, {and} \bibinfo{person}{Geehyuk Lee}.} \bibinfo{year}{2020}\natexlab{}.
\newblock \showarticletitle{Speech Gesture Generation from the Trimodal Context of Text, Audio, and Speaker Identity}.
\newblock \bibinfo{journal}{\emph{ACM Trans. Graph.}} \bibinfo{volume}{39}, \bibinfo{number}{6}, Article \bibinfo{articleno}{222} (\bibinfo{date}{nov} \bibinfo{year}{2020}), \bibinfo{numpages}{16}~pages.
\newblock
\showISSN{0730-0301}
\urldef\tempurl%
\url{https://doi.org/10.1145/3414685.3417838}
\showDOI{\tempurl}


\end{thebibliography}

\appendix

\end{document}